%% file: robotgps.tex
\documentclass[letterpaper, 10 pt, conference]{ieeeconf}  

\IEEEoverridecommandlockouts                              

\usepackage{url}
\usepackage{amsmath}
\usepackage{graphicx}
\usepackage{verbatim}
\usepackage{algorithm}
\usepackage{algorithmic}
\usepackage{hyperref}
\usepackage{color}
\usepackage{wrapfig}
\usepackage[small]{caption}

\include{defs}

\title{\LARGE \bf
Learning Contact-Rich Manipulation Skills with Guided Policy Search
}

\author{Sergey Levine, Nolan Wagener, Pieter Abbeel
\thanks{Department of Electrical Engineering and Computer Science, University of California, Berkeley, Berkeley, CA 94709}%
}

\begin{document}

\maketitle
\thispagestyle{empty}
\pagestyle{empty}

\begin{abstract}

Autonomous learning of object manipulation skills can enable robots to acquire rich behavioral repertoires that scale to the variety of objects found in the real world. However, current motion skill learning methods typically restrict the behavior to a compact, low-dimensional representation, limiting its expressiveness and generality. In this paper, we extend a recently developed policy search method \cite{la-lnnpg-14} and use it to learn a range of dynamic manipulation behaviors with highly general policy representations, without using known models or example demonstrations. Our approach learns a set of trajectories for the desired motion skill by using iteratively refitted time-varying linear models, and then unifies these trajectories into a single control policy that can generalize to new situations. To enable this method to run on a real robot, we introduce several improvements that reduce the sample count and automate parameter selection. We show that our method can acquire fast, fluent behaviors after only minutes of interaction time, and can learn robust controllers for complex tasks, including putting together a toy airplane, stacking tight-fitting lego blocks, placing wooden rings onto tight-fitting pegs, inserting a shoe tree into a shoe, and screwing bottle caps onto bottles.

\end{abstract}

\section{Introduction}

Autonomous acquisition of manipulation skills has the potential to dramatically improve both the ease of deployment of robotic platforms, in domains ranging from manufacturing to household robotics, and the fluency and speed of the robot's motion. It is often much easier to specify \emph{what} a robot should do, by means of a compact cost function, than \emph{how} it should do it, and manipulation skills that are learned from real-world experience can leverage the real dynamics of the robot and its environment to accomplish the task faster and more efficiently. For tasks with a significant force or dynamics component, such as inserting a tight-fitting part in an assembly task, standard kinematic methods either require very high precision or fail altogether, while methods that learn from real-world interaction could in principle navigate the complex dynamics of the task without prior knowledge of the object's physical properties.

Although significant progress has been made in this area in recent years \cite{kbp-rlrs-13,dnp-spsr-13}, learning robot motion skills remains a major challenge. Policy search is often the method of choice due to its ability to scale gracefully with system dimensionality \cite{dnp-spsr-13}, but successful applications of policy search typically rely on using a compact, low-dimensional representation that exposes a small number of parameters for learning \cite{kp-pgrlf-04,tzs-spgrl-04,gpw-fbwrc-06,emmnc-lcbbl-08,ps-rlmsp-08,phas-lgmsl-09}. Substantial improvements on real-world systems have come from specialized and innovative policy classes \cite{ins-lallm-03}, and designing the right low-dimensional representation often poses a significant challenge.


\begin{figure}
\setlength{\unitlength}{0.5\columnwidth}
\begin{picture}(1.99,0.93) \linethickness{0.5pt}

\put(0.075,-0.04){\includegraphics[width=0.9\columnwidth]{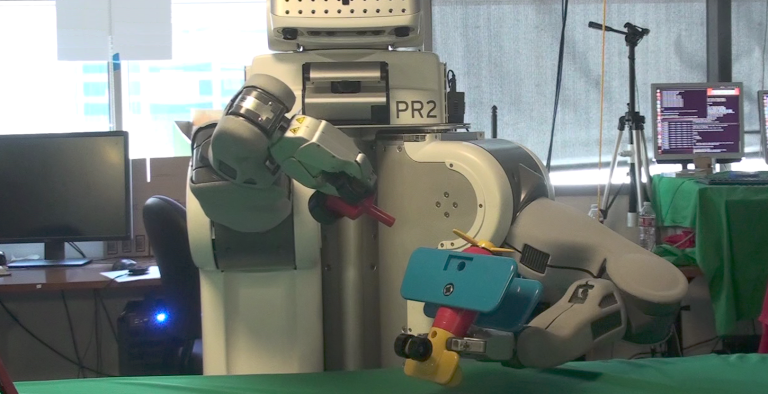}}

\end{picture}
\caption{PR2 learning to attach the wheels of a toy airplane.
\label{fig:teaser}
}
\vspace{-0.2in}
\end{figure}

In this paper, we show that a range of motion skills can be learned using only general-purpose policy representations. We use our recently developed policy search algorithm \cite{la-lnnpg-14}, which combines a sample-efficient method for learning linear-Gaussian controllers with the framework of guided policy search, which allows multiple linear-Gaussian controllers (trained, for example, from several initial states, or under different conditions) to be used to train a single nonlinear policy with any parameterization, including complex, high-dimensional policies represented by large neural networks. This policy can then generalize to a wider range of conditions than the individual linear-Gaussian controllers.

We present several modifications to this method that make it practical for deployment on a robotic platform. We introduce an adaptive sample count adjustment scheme that minimizes the amount of required system interaction time, and we develop a step size adaptation method that allows our algorithm to learn more aggressively in easier stages of the task, further reducing the number of real-world samples. To handle the resulting scarcity of training data in the guided policy search procedure, we also propose a method for augmenting the training set for the high-dimensional nonlinear policy using synthetic samples, and show that this approach can learn complex neural network policies from a small number of real-world trials. Finally, we propose a general framework for specifying cost functions for a broad class of manipulation skills, which is used in all of our experiments. Our experimental results include putting together a toy airplane (Figure~\ref{fig:teaser}), stacking tight-fitting lego blocks, placing wooden rings onto tight-fitting pegs, inserting a shoe tree into a shoe, and screwing bottle caps onto bottles.

\section{Related Work}

Direct policy search is a promising technique for learning robotic motion skills, due to its ability to scale to high-dimensional, continuous systems. Policy search has been used to train controllers for games such as ball-in-cup and table tennis \cite{kop-rlarm-10}, manipulation \cite{phas-lgmsl-09,drf-lclcm-11}, and robotic locomotion \cite{kp-pgrlf-04,tzs-spgrl-04,gpw-fbwrc-06,emmnc-lcbbl-08}. While contact-rich manipulation tasks such as the ones considered in this work are less common, several prior methods have addressed opening doors and picking up objects \cite{krps-lfcpc-11}, as well as using force sensors during manipulation \cite{pkcts-sltop-11}. Other reinforcement learning techniques, such as dynamic programming and temporal difference learning, have also been employed, though the curse of dimensionality has typically prevented them from being used to directly control high-dimensional articulated robots. An overview of recent reinforcement learning methods in robotics can be found in a recent survey paper \cite{kbp-rlrs-13}.

Previous policy search methods in robotics generally use one or more of the following techniques to achieve good performance and reasonable training times: the use of example demonstrations \cite{ghcb-rlicr-07,phas-lgmsl-09,tbs-rlmsh-10}, the use of a prior controller that allows the policy to only set high level targets, such as kinematic end-effector motion \cite{phas-lgmsl-09}, and the use of carefully designed policy classes that ensure that most parameter settings produce reasonable behavior \cite{kp-pgrlf-04,tzs-spgrl-04,gpw-fbwrc-06,emmnc-lcbbl-08}. In contrast, our method learns general-purpose time-varying linear-Gaussian controllers, which can represent any Gaussian trajectory distribution and contain thousands of parameters. These controllers can further be used within the framework of guided policy search to learn a policy with any parameterization, including high-dimensional policies such as large neural networks. Although a number of methods have been used to train neural network policies \cite{hr-nrlcr-07,ggb-lrac-92}, prior applications have required large amounts of training time and have been restricted either to low-dimensional systems or to high-level control. We show that our approach can quickly train neural networks with over 4000 parameters for a variety of complex manipulation tasks, directly controlling the torques at every joint of a 7 DoF manipulator arm. Furthermore, although our method could easily be initialized with demonstrations, we show in our experiments that it can quickly learn manipulation behaviors completely from scratch, simultaneously discovering how to actuate the robot's joints and how to accomplish the task.

Our algorithm is based on our recent work on learning policies with trajectory optimization under unknown dynamics \cite{la-lnnpg-14}, which has been shown to learn complex policies under unknown system dynamics even with a modest amount of system interaction time. Previous guided policy search algorithms have also focused on leveraging known models of the environment \cite{lk-gps-13,lk-vpsto-13,lk-lcnnp-14} and more sophisticated trajectory optimization algorithms \cite{mt-cbfat-14}. Our linear-Gaussian controller representation is related to other trajectory-centric representations in policy search, such as dynamic movement primitives (DMPs) \cite{ins-lallm-03}. However, while DMPs are primarily kinematic, linear-Gaussian controllers explicitly encode a distribution over actions (torques) in terms of the robot's state at each time step. Optimization of these linear-Gaussian controllers can be viewed as accomplishing a similar task to iterative learning control (ILC), though ILC optimizes a controller for following a known trajectory \cite{bta-silc-06}, while our approach also discovers the trajectory itself. In this light, the guided policy search (GPS) component of our method can be viewed as leveraging the strengths of trajectory-centric controller optimization to learn a more complex policy that can generalize to new situations. Simply using a set of trajectories as a training set for supervised learning is however insufficient \cite{rgb-rilsp-11}. Instead, GPS iteratively adapts the trajectories to match the policy, eventually bringing their state distributions into agreement.


\section{Overview}

We formulate robotic motion skill learning as policy search, where the aim is to find a good setting of the parameters $\params$ for a policy $\policy_\params(\action_t|\state_t)$, which specifies a probability distribution over actions at each time step, conditioned on the state. In a robotic system, the actions might be joint torques, and the state might correspond to the configuration of the robot (joint angles and velocities) and the environment (e.g. the position of a target object). The policy parameters are optimized with respect to the expected cost of the policy, given by $E_{\policy_\params}[\sum_{t=1}^T\cost(\st,\at)]$. The expectation is taken under the policy's trajectory distribution $p(\traj)$, which consists of the policy $\policy_\params(\at|\st)$ and the dynamics $p(\state_{t+1}|\st,\at)$. We abbreviate this expectation as $E_{\policy_\params}[\cost(\traj)]$ for convenience.

Our method consists of two parts: an algorithm that trains time-varying linear-Gaussian controllers of the form $\trajdist(\at|\st) = \gauss(\Kpol_t\st + \kpol_t,\ucovar_t)$, which can be thought of as optimizing a trajectory together with stabilizing feedbacks $\Kpol_t$, and a guided policy search component that combines one or more such time-varying linear-Gaussian controllers to learn a nonlinear policy with an arbitrary parameterization. For clarity, we will reserve the term policy and the symbol $\policy_\params$ for this final, nonlinear policy, while the linear-Gaussian policy will be referred to as a controller, denoted by $\trajdist$.

The advantage of this approach over directly learning $\policy_\params$ is that the optimization of the linear-Gaussian controllers can exploit their special structure to learn very quickly with only a small amount of system interaction, while the nonlinear policy is optimized with supervised learning to match the linear-Gaussian controllers, alleviating many of the challenges associated with learning complex, high-dimensional policies in standard reinforcement learning methods.

Both the linear-Gaussian controller and the nonlinear policy are highly general representations. The linear-Gaussian controller can represent any trajectory, and by including additional variables in the state (such as the vector to the target in a reaching task), it can perform feedback on a rich set of features. As we show in Section~\ref{sec:trajresults}, the linear-Gaussian controller alone can be sufficient for a range of tasks, and provides good robustness to small perturbations. When the initial conditions can vary drastically, using multiple linear-Gaussian controllers to train a complex nonlinear policy, represented, for instance, by a large neural network, can provide sufficient generalization to succeed in new, previously unseen situations, as shown in Section~\ref{sec:polresults}. This representation is even more general and, with enough hidden units, can capture any stationary policy.

\section{Trajectory Optimization under Unknown Dynamics}
\label{sec:to}

Policy search methods can be divided into model-based methods, which learn a model of the system and optimize the policy under this model \cite{dnp-spsr-13}, and model-free methods, which improve the policy directly, for example by estimating its gradient using samples \cite{ps-rlmsp-08}. Model-based methods must learn a global model of the system, which can be extremely challenging for complex manipulation tasks, while model-free methods tend to require a large amount of interaction time. To optimize the linear-Gaussian controllers $\trajdist(\at|\st)$, we employ a hybrid approach that learns local models around the current trajectory, combining the flexibility of model-free techniques with the efficiency of model-based methods. While prior policy search methods have been proposed that leverage locally linear models (see, e.g., \cite{lpnp-sbits-14}), the LQR-based update discussed in this section has been previously shown to achieve substantially better sample efficiency on challenging, contact-rich problems \cite{la-lnnpg-14}.

We use the samples collected under the last controller to fit time-varying linear-Gaussian dynamics $p(\state_{t+1}|\st,\at) = \gauss(\fxt\st + \fut\at,\noise_t)$, and then solve a variant of the linear-quadratic-Gaussian (LQG) problem to find a new controller $\trajdist(\at|\st)$ that is optimal under this model, subject to the constraint that its trajectory distribution $\trajdist(\traj)=\prod_t p(\state_{t+1}|\st,\at)p(\at|\st)$ deviates from the previous one $\hat{\trajdist}(\traj)$ by a bounded amount. This bound ensures that the linear model remains mostly valid for the new controller. The derivation in Sections~\ref{sec:kldivlqg} and \ref{sec:bgdistro} follows our prior work \cite{la-lnnpg-14}, and we discuss some novel improvements Section~\ref{sec:adaptive}.

\subsection{KL-Divergence Constrained LQG}
\label{sec:kldivlqg}

In the LQG setting, the optimal feedback controller can be computed by a backward dynamic programming algorithm that computes the $Q$-function and value function at each time step, starting at the end of the trajectory. These functions are quadratic, and are given by the following recurrence:
\begin{align*}
\Qyyt &= \costhesst + \fyt\tr\Vxxtp\fyt \\
\Qyt &= \costgradt + \fyt\tr\Vxtp \\
\Vxxt &= \Qxxt - \Quxt\tr\Quut\inv\Qux \\
\Vxt &= \Qxt - \Quxt\tr\Quut\inv\Qut.
\end{align*}
Subscripts denote derivatives: $\costgradt$ is the derivative of the cost at time $t$ with respect to $(\st,\at)\tr$, $\costhesst$ is the Hessian, and so forth. The optimal controller is given by $\detpolicy(\state_t) = \haction_t + \kpol_t + \Kpol_t(\state_t - \hstate_t)$, where $\Kpol_t = -\Quut\inv \Quxt$ and $\kpol_t = -\Quut\inv \Qut$. We can also show that the linear-Gaussian controller \mbox{$\trajdist(\at|\st) = \gauss(\bar{\action}_t + \kpol_t + \Kpol_t(\state_t - \hstate_t),\Quut\inv)$} is the optimal solution to the \emph{maximum entropy} objective $\min E_{\trajdist}[\cost(\traj)] - \ent(\trajdist(\traj))$, where the usual minimum cost is augmented with an entropy maximization term \cite{lk-gps-13}.

To update the linear-Gaussian controller $\trajdist(\at|\st)$ without deviating too far from the previous trajectory distribution $\bar{\trajdist}(\traj)$, we must optimize a constrained objective, given by
\[
\min_{\trajdist(\traj)\in\gauss(\traj)} E_{\trajdist}[\cost(\traj)] \text{ s.t. } \kl(\trajdist(\traj)\|\hat{\trajdist}(\traj)) \leq \epsilon.
\]
\noindent This type of policy update has previously been proposed by several authors in the context of policy search \cite{bs-cps-03,ps-rlmsp-08,pma-reps-10}. However, the special structure of the linear-Gaussian controller and dynamics admits a particularly simple solution in our case. The Lagrangian of this problem is
\begin{align*}
&\lagrangian(\trajdist,\eta) = E_{\trajdist}[\cost(\traj)] + \eta [\kl(\trajdist(\traj)\|\hat{\trajdist}(\traj)) - \epsilon] = \\
&\left[  \sum_t\! E_{\trajdist(\state_t,\action_t)}[\cost(\st,\at) \!-\! \eta\log\hat{\trajdist}(\at|\st)] \right] \!-\! \eta\ent(\trajdist(\traj)) \!-\! \eta\epsilon,
\end{align*}
\noindent where the simplification results from assuming that the two distributions share the same dynamics. This Lagrangian has the same form as the maximum entropy objective when the cost is set to \mbox{$\tilde{\cost}(\state_t,\action_t) = \frac{1}{\eta}\cost(\state_t,\action_t) - \log\hat{\trajdist}(\action_t|\state_t)$}. Therefore, we can minimize it with respect to the primal variables (the parameters of $\trajdist(\at|\st)$) by solving the LQG problem under the modified cost $\tilde{\cost}$. The dual variable $\eta$ is obtained with dual gradient descent (DGD) \cite{b-co-04}, by repeatedly solving the LQG problem and updating $\eta$ by the amount of constraint violation. A bracket line search on $\eta$ can reduce the required number of DGD iterations to only 3 to 5.

Unfortunately, the sample complexity of fitting the linear dynamics scales with the dimensionality of the system. In the following sections, we discuss several improvements that can greatly reduce the required number of samples, making the method practical for real robotic systems.

\subsection{Background Dynamics Distributions}
\label{sec:bgdistro}

For high-dimensional systems, gathering enough samples to fit the linear dynamics can result in very long training times. We can greatly reduce the required number of samples by using a simple insight: the dynamics at nearby time steps and prior iterations are strongly correlated with the dynamics at a particular time in the current iteration. We can therefore use all of these additional dynamics samples to construct a prior for fitting $\gauss(\fxt\st + \fut\at,\noise_t)$ at each step $t$.

A good choice for this prior is a Gaussian mixture model (GMM). As discussed in prior work \cite{kb-ialsn-10}, GMMs are well suited for modeling piecewise linear dynamics, which are a good approximation for articulated systems (such as robots) that contact objects in the environment. We build a GMM at each iteration from points $(\st,\at,\state_{t+1})\tr$, gathered from all steps at the current iteration and several prior iterations. In each cluster $\cluster_i$, the conditional distribution $\cluster_i(\state_{t+1}|\st,\at)$ is a linear-Gaussian dynamics model, while the marginal $\cluster_i(\st,\at)$ is the region where this model is valid.

To use the GMM as a prior on the linear dynamics, we find the average cluster weights of the samples at each step and compute their weighted mean and covariance. This mean and covariance then serves as an normal-inverse-Wishart prior for fitting a Gaussian to $(\st,\at,\state_{t+1})\tr$ at step $t$, and the dynamics are obtained by conditioning on $(\st,\at)\tr$.

We previously found that this GMM allows us to take many fewer samples than there are state dimensions \cite{la-lnnpg-14}. In this work, we reduce the number of samples even further by using an adaptive sample count adjustment scheme, and we reduce the number of iterations (and therefore total samples) by adaptively inreasing the step size $\epsilon$, as described below.

\subsection{Adaptive Adjustment of Step Size and Sample Count}
\label{sec:adaptive}

The limit $\epsilon$ on the KL-divergence between the new and old trajectory distributions acts as a step size: large values can speed up learning, but at the risk of taking steps that are too large and do not improve the objective. In our prior work, we heuristically decreased $\epsilon$ when the method failed to improve the objective \cite{la-lnnpg-14}. Here, we introduce a more sophisticated step size adjustment method that both increases and decreases the step size, by explicitly modeling the additional cost at each iteration due to unmodeled changes in the dynamics. In our experiments, we found that this scheme removed the need to manually tune the step size, and achieved significantly faster overall learning.

After taking samples from the current controller and fitting its dynamics, we estimate the step size $\epsilon$ that would have been optimal at the previous iteration by comparing the expected decrease in cost under the previous dynamics to the actual cost under the current ones. Let $\cost_{k-1}^{k-1}$ denote the cost under the previous dynamics and previous controller, $\cost_{k-1}^{k}$ under the previous dynamics and current controller, and $\cost_k^k$ under the current dynamics and controller. We assume that the improvement in cost is linear in $\epsilon$ and the additional cost due to unmodeled changes in the dynamics is quadratic:
\[
\cost_k^k - \cost_{k-1}^{k-1} = \Penalty \epsilon^2 + \Rate \epsilon,
\]
\noindent where $\Penalty$ is the additional cost due to unmodeled changes in the dynamics and $\Rate = (\cost_{k-1}^k - \cost_{k-1}^{k-1})/ \epsilon$ is the expected linear improvement rate. Since we know all values except $\Penalty$, we can solve for $\Penalty$. To pick a new step size $\epsilon^\prime$, we use the step size that would have been optimal at the previous iteration based on the known values of $\Penalty$ and $\Rate$:
\[
\epsilon^\prime = -\frac{\Rate}{2\Penalty} = \frac{1}{2}\epsilon (\cost_{k-1}^k - \cost_{k-1}^{k-1})/(\cost_{k-1}^k - \cost_k^k).
\]
To compute $\cost_{k-1}^{k-1}$, $\cost_{k-1}^k$, and $\cost_k^k$, we note that under linear dynamics, the marginals $\trajdist(\st,\at)$ are Gaussian (see e.g. \cite{lk-vpsto-13}), and we can evaluate $E_\trajdist[\cost(\traj)] = \sum_{t=1}^T E_{\trajdist(\st,\at)}[\cost(\st,\at)]$ analytically when using a local quadratic expansion of the cost. This also justifies our assumption that cost change is linear in the KL-divergence, since the KL-divergence is quadratic in the mean and linear in the covariance, as is the expectation of the quadratic cost.

In addition to automatically setting the step size, we can also use the estimated expected cost to adjust the number of samples to take at the next iteration. We employ a simple heuristic that compares the prediction of the expected cost under the estimated dynamics with the Monte Carlo estimate obtained by averaging the actual cost of the samples. The intuition is that, if there are too few samples to accurately estimate the dynamics, the expected cost under these dynamics will differ from the Monte Carlo estimate. We increase the sample count by one if the analytic estimate falls more than one standard deviation outside of the Monte Carlo mean, and decrease it if it falls within half a standard deviation.

\section{General Parameterized Policies}
\label{sec:gps}

While the approach in the previous section can solve a range of tasks, as shown in Section~\ref{sec:trajresults}, it only provides robustness against small variations in the initial conditions. This is not a problem in some cases, such as when the robot is already gripping both of the objects that it must manipulate (and therefore can position them in the correct initial state), but we often want to learn policies that can generalize to many initial states. Previous work has addressed this by explicitly defining the policies in terms of object positions, effectively hard-coding how generalization should be done \cite{phas-lgmsl-09}. We instead train the policy on multiple initial conditions. For example, to attach one lego block to another at any location, we might train on four different locations.

\subsection{Guided Policy Search}

Simple linear-Gaussian controllers typically cannot handle such variation, so we must train a more expressive parameterized policy. Policies with high-dimensional parameterizations are known to be a major challenge for policy search techniques \cite{dnp-spsr-13}, but we can still leverage the linear-Gaussian method from the preceding section by using the framework of guided policy search (GPS) \cite{lk-gps-13,lk-vpsto-13,lk-lcnnp-14}. With GPS, the final policy is not trained directly with reinforcement learning. Instead, a set of trajectories are optimized, using for example the method in the previous section, and samples from these trajectories are used as the training set for supervised training of the policy. Since supervised learning can reliably optimize function approximators with thousands of parameters, we can use GPS with very rich policy classes.

Training policies with supervised learning is not guaranteed to improve their expected cost, since the state distribution of the training set does not match that of the policy \cite{rgb-rilsp-11}. In the constrained guided policy search algorithm, which we use in this work, this is addressed by reoptimizing the trajectories to better match the current policy, such that their state distributions match at convergence \cite{lk-lcnnp-14}. This approach aims to solve the following constrained optimization problem:
\begin{align*}
\min_{\params,\trajdist(\traj)} E_{\trajdist(\traj)}[\cost(\traj)] \text{ s.t. } \kl(\trajdist(\state_t)\policy_\params(\action_t|\state_t)\|\trajdist(\state_t,\action_t)) = 0 \,\,\forall t,
\end{align*}
\noindent where $\trajdist(\traj)$ is a trajectory distribution, and $\policy_\params$ is the parameterized policy.\footnote{In the case of multiple training conditions (e.g. target positions), each with their own separate trajectory $\trajdist(\traj)$, the constraint and objective are replicated for each trajectory, and the policy is shared between all of them.} When the constraint is satisfied, $\policy_\params$ and $\trajdist(\traj)$ have the same state distribution, making supervised learning a suitable way to train the policy. The constrained problem is solved by relaxing the constraint and applying dual gradient descent (note that this instance of DGD is not related to the use of DGD in the preceding section). The Lagrangian is
\begin{align*}
&\lagrangiangps(\params,\trajdist,\polwt) = \\
&E_{\trajdist(\traj)}[\cost(\traj)] + \sum_{t=1}^T \polwt_t \kl(\trajdist(\state_t)\policy_\params(\action_t|\state_t)\|\trajdist(\state_t,\action_t)).
\end{align*}
As is usual with DGD, we alternate between optimizing the Lagrangian with respect to the primal variables (the trajectory distributions and policy parameters), and taking a subgradient step on the dual variables $\polwt_t$. The primal optimization is itself done in alternating fashion, alternating between the policy and the trajectories. The policy optimization corresponds to supervised learning, weighted by the Q-function of the trajectories, and the trajectory optimization is performed as described in the preceding section. The policy KL-divergence term in the trajectory objective requires a slightly different dynamic programming pass, which is described in prior work \cite{lk-lcnnp-14,la-lnnpg-14}. The alternating optimization is done for only a few iterations before each dual variable update, rather than to convergence, which greatly speeds up the method. Pseudocode is provided in Algorithm~\ref{alg:gps}.

\begin{algorithm}[tb]
\caption{Guided policy search with unknown dynamics}
\label{alg:gps}
\begin{algorithmic}[1]
\FOR{iteration $k=1$ to $K$}
\STATE Generate samples $\{\traj_i^j\}$ from each linear Gaussian controller $\trajdist_i(\traj)$ by running it on the robot
\STATE Minimize $\sum_{i,t}\! \polwt_{i,t}\kl(\trajdist_i(\state_t)\policy_\params(\action_t|\state_t)\|\trajdist_i(\state_t,\action_t))$ with respect to $\params$ using samples $\{\traj_i^j\}$
\STATE Update $\trajdist_i(\at|\st)$ using the LQG-like method
\STATE Increment each of the dual variables $\polwt_{i,t}$ by $\alpha \kl(\trajdist_i(\state_t)\policy_\params(\action_t|\state_t)\|\trajdist_i(\state_t,\action_t))$
\ENDFOR
\STATE {\bf return} optimized policy parameters $\params$
\end{algorithmic}
\end{algorithm}

\subsection{Augmenting Policy Training with Synthetic Samples}

We previously discussed a number of improvements that allow us to reduce the number of samples during training. However, one effect of using so few samples is that the training set for the nonlinear policy becomes very small, causing the policy to overfit to the current samples.

We alleviate this issue by artificially augmenting the training set for the policy with synthetic samples. Since the policy is simply trained on pairs of states and actions, where the states are drawn from the state distribution of each trajectory distribution, there is no need to get these samples from the real system. We can form the state marginals $\trajdist(\st)$ as described in Section~\ref{sec:adaptive}, sample the desired number of states, and compute the mean action at each state under the linear-Gaussian controller. In this way, we can construct an unlimited training set. In practice, we generate 50 such synthetic samples from each trajectory at each time step.


\section{Defining Objectives for Robotic Manipulation}
\label{sec:cost}

As with all policy search methods, we require the desired task to be defined by a cost function $\cost(\st,\at)$. In practice, the cost function both defines the task, and provides general guidance about the directions in state space that improve task performance (this is sometimes referred to as cost shaping). Choosing a good cost is important for minimizing system interaction time, but it is also important to design the cost such that it does not require manual engineering or tuning for each behavior. We propose a general approach for constructing cost functions for manipulation tasks that involve positioning a grasped object, either in space or in relation to another object. This task comes up often in industrial and household robotics, and includes assembly (placing a part into its intended slot), working with electronic devices (inserting a plug into a socket), and various other tasks. Note that the object is usually not positioned in free space, and placing it into the right configuration often involves a complex dynamic operation with numerous frictional contacts.

We define our cost in terms of the desired position of several points on the object. The easiest way to define these positions is to manually position the object as desired. In our experiments, we positioned the robot's gripper over the object to determine its location, though any vision system (such as Kinect-based point cloud registration) could also perform this step. Note that we do not require a demonstration of \emph{how} the robot should perform the task, only of the final position of the object, which can be positioned by a human user. For more complex tasks, such as the shoe tree described in the next section, we can also specify an intermediate waypoint to go around obstacles. On the bottle cap tasks, we added a term to encourage the wrist to spin at a set rate, encoding the prior knowledge that the cap should turn.

At each time step, the cost depends on the distance between the points on the object $\points_t$ (where all points are concatenated into a single vector), and the points at the target location $\points^\star$, and is given by some penalty $\costnorm(\vnorm{\points_t-\points_t^\star})$. The shape of $\costnorm$ has a large effect on the final behavior. The quadratic penalty $\costnorm(d) = d^2$ is generally inadequate, since it does not sufficiently penalize small errors in position, while the tasks typically require very precise positioning: when assembling a plastic toy, it is not enough to throw the parts together, they must properly slot into the correct fittings. We therefore employ a more complex penalty function given by the following equation:
\[
\costnorm(d) = w d^2 + v \log(d^2 + \alpha).
\]
\begin{wrapfigure}{r}{.3\columnwidth}
\vspace{-0.2in}
\hspace{-0.02\columnwidth}\includegraphics[height=0.29\columnwidth]{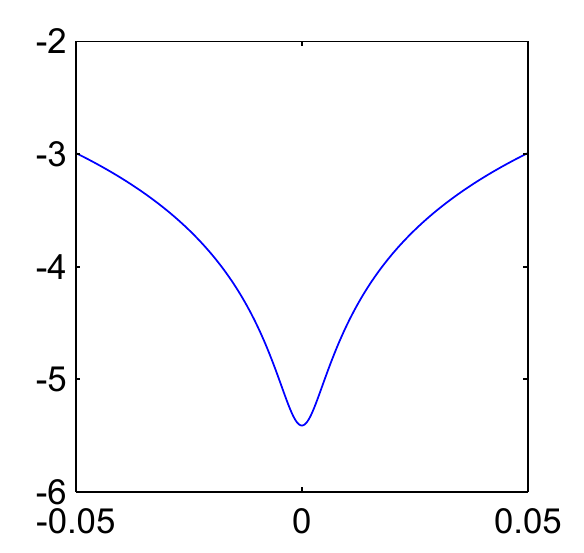}
\vspace{-0.3in}
\end{wrapfigure}
The squared distance term encourages the controller to quickly get the object in the vicinity of the target, while the second term (sometimes called the Lorentzian $\rho$-function) has a concave shape that encourages precise placement at the target position. An illustration of this function is shown above. The depth of the ``funnel'' and the width of its floor are determined by $\alpha$, while $w$ and $v$ trade off the two penalties. We set $w = 1$, $v = 1$, $\alpha = 10^{-5}$. The value of $\alpha$ depends on the desired distance to the target, and can be chosen by examining the shape of $\log(d + \alpha)$.\footnote{An automatic scheme could also be devised by solving for the value of $\alpha$ that sets the inflection point of $\log(d + \alpha)$ at the desired tolerance.} In our trajectory optimization algorithm, the negative curvature of $\costnorm$ encourages the algorithm to converge on the target more quickly and more precisely at time steps that are already close to the target. In addition to this penalty, we quadratically penalize joint velocities and torques to create smooth, controlled motions. We found that this cost function worked well for all of the manipulation tasks we evaluated, and the ease of specifying the target makes it an appealing option for a user-friendly learning framework.





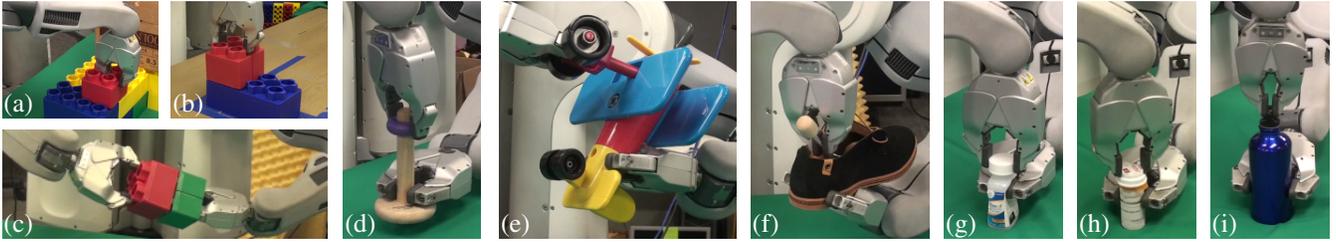
\begin{figure*}
\setlength{\unitlength}{0.5\columnwidth}
\begin{picture}(1.99,0.75) \linethickness{0.5pt}

\put(0.0,0.37){\includegraphics[width=0.24\columnwidth]{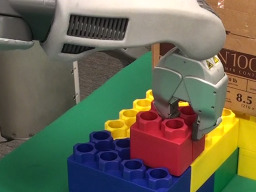}}
\put(0.52,0.37){\includegraphics[width=0.24\columnwidth]{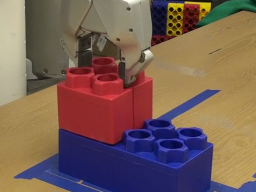}}
\put(0.0,0.0){\includegraphics[width=0.50\columnwidth]{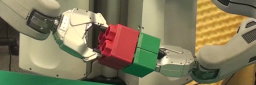}}

\put(1.05,0.0){\includegraphics[height=0.365\columnwidth]{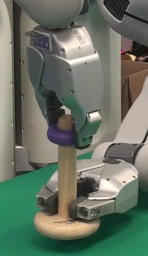}}
\put(1.53,0.0){\includegraphics[height=0.365\columnwidth]{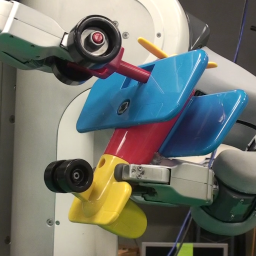}}
\put(2.305,0.0){\includegraphics[height=0.365\columnwidth]{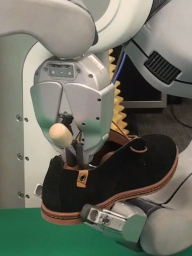}}

\put(2.90,0.0){\includegraphics[height=0.365\columnwidth]{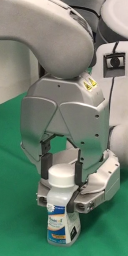}}
\put(3.31,0.0){\includegraphics[height=0.365\columnwidth]{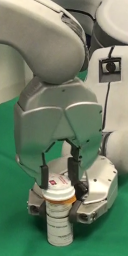}}
\put(3.72,0.0){\includegraphics[height=0.365\columnwidth]{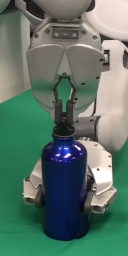}}

\put(0.005,0.39){\textcolor{white}{(a)}}
\put(0.525,0.39){\textcolor{white}{(b)}}
\put(0.005,0.02){\textcolor{white}{(c)}}
\put(1.055,0.02){\textcolor{white}{(d)}}
\put(1.535,0.02){\textcolor{white}{(e)}}
\put(2.31,0.02){\textcolor{white}{(f)}}
\put(2.905,0.02){\textcolor{white}{(g)}}
\put(3.315,0.02){\textcolor{white}{(h)}}
\put(3.725,0.02){\textcolor{white}{(i)}}

\end{picture}
\caption{Tasks in our experiments: (a) stacking large lego blocks on a fixed base, (b) onto a free-standing block, (c) held in both gripper; (d) threading wooden rings onto a tight-fitting peg; (e) assembling a toy airplane by inserting the wheels into a slot; (f) inserting a shoe tree into a shoe; (g,h) screwing caps onto pill bottles and (i) onto a water bottle. Videos are included with the supplementary material and at \url{http://rll.berkeley.edu/icra2015gps/index.htm}.
\label{fig:objects}
\vspace{-0.2in}
}
\end{figure*}

\section{Experimental Results}
\label{sec:experiments}

We conducted a set of experiments using both the linear-Gaussian training procedure in isolation, and in combination with guided policy search. In the latter case, the final nonlinear policy was represented by a neural network with two hidden layers. We chose tasks that involve complex dynamics: stacking tight-fitting lego blocks, assembling plastic toys, inserting a shoe tree into a shoe, placing tight-fitting wooden rings onto pegs, and screwing bottle caps onto bottles. The lego blocks were tested in three conditions: attaching a block to a heavy base that is fixed to the table, attaching a block to another free-standing block (which can shift due to the forces applied during stacking), and attaching two blocks that are both held by the robot. These tasks are especially challenging for standard motion planning methods, which usually deal only with kinematics. Dynamic methods (such as model-predictive control) would also find these tasks difficult, since the physical properties of the objects are not known in advance and are often difficult to model. Images of the objects in our experiments are presented in Figure~\ref{fig:objects}.

In all experiments, both the linear-Gaussian controllers and neural networks directly commanded the torque on each of the seven joints on the robot's arm at 20 Hz, and took as input the current joint angles and velocities, the Cartesian velocities of two or three points on the manipulated object (two for radially symmetric objects like the ring, three for all others), the vector from the target positions of these points to their current position, and the torque applied at the previous time step. The object was assumed to be rigidly attached to the end-effector, so that forward kinematics could be used to compute the object's current position.


\subsection{Linear-Gaussian Controllers}
\label{sec:trajresults}

We first trained controllers for each of the manipulation tasks using only the linear-Gaussian method in Section~\ref{sec:to}. While this approach has limited ability to generalize to different initial states, certain manipulation scenarios are less vulnerable to this limitation. For example, when the objects being manipulated are already grasped by the robot, they can be positioned in the suitable initial state automatically, allowing even these simple controllers to succeed. Furthermore, although the controllers are linear-Gaussian, since the dynamics are learned, feedback can be performed on any observation signal, not just the state of the robot. As described in the previous paragraph, we include the position of points on the object (assumed to be in the frame of the end-effector) to improve robustness to small variations.

\begin{figure}[t]
\setlength{\unitlength}{0.5\columnwidth}
\begin{picture}(1.99,0.92) \linethickness{0.5pt}

\put(0.0,0.0){\includegraphics[height=0.45\columnwidth]{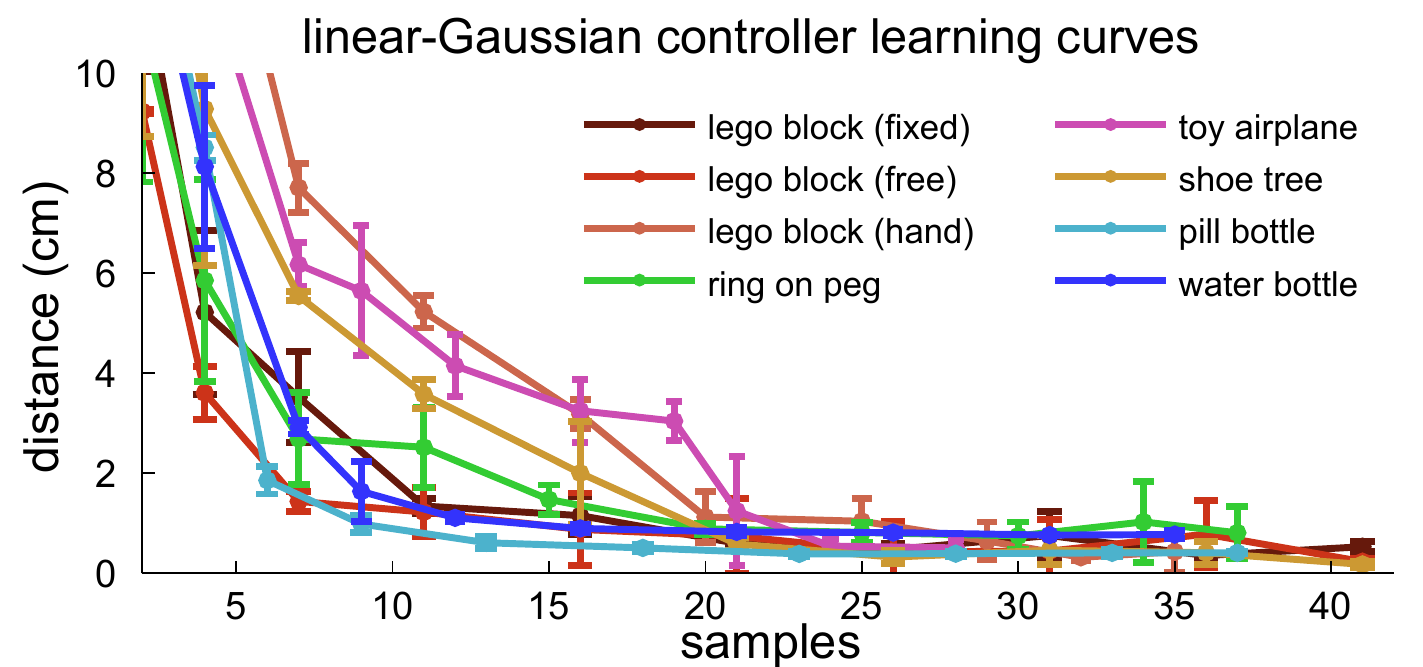}}

\end{picture}
\caption{Distance to specified target point per iteration during training of linear-Gaussian controllers. The actual target location may differ due to perturbations. Error bars indicate one standard deviation. Note that the samples per iteration vary slightly due to adaptive sample count selection.
\label{fig:lgresults}
\vspace{-0.2in}
}
\end{figure}

In Figure~\ref{fig:lgresults}, we show learning curves for each of the tasks. The curves are shown in terms of the number of samples, which changes between iterations due to the adaptive sampling rule in Section~\ref{sec:adaptive}. Note that the number of samples required to learn a successful controller is in the range of 20-25, substantially lower than many previously proposed policy search methods in the literature \cite{ps-rlmsp-08,kop-rlarm-10,tbs-rlmsh-10,dnp-spsr-13}. Total learning time was about ten minutes for each task, of which only 3-4 minutes involved system interaction (the rest included resetting to the initial state and computation time, neither of which were optimized).

The supplementary video\footnote{See \url{http://rll.berkeley.edu/icra2015gps/index.htm}} shows each of the controllers performing their task. A few interesting strategies can be observed. In the toy airplane task, the round peg on the wheels is inserted first, and the motion of the peg against the slot is used to align the gear base for the final insertion. Note how the wheels are first pulled out of the slot, and then inserted back in at the right angle. For the ring task, the controller spends extra time positioning the ring on top of the peg before applying downward pressure, in order to correct misalignments. Friction between the wooden ring and the peg is very high, so the alignment must be good. The shoe tree task used an extra waypoint supplied by the user to avoid placing the shoe tree on top of the shoe, but careful control is still required after the waypoint to slide the front of the shoe tree into the shoe without snagging on the sides. For the bottle task, we tested a controller trained on one pill bottle on a different pill bottle (see Figure~\ref{fig:objects}), and found that the controller could still successfully complete the task.

To systematically evaluate the robustness of the controllers, we conducted experiments on the lego block and ring tasks where the target object (the lower block and the peg) was perturbed at each trial during training, and then tested with varying amounts of perturbation. For each task, controllers were trained with Gaussian perturbations with standard deviations of $0$, $1$, and $2$ cm in the position of the target object. For the lego block, the perturbation was applied to the corners, resulting in a rotation. A different perturbation was sampled at every trial, and the controller was unaware of the noise, always assuming the object to be at the original position. These controllers were then tested on perturbations with a radius of $0$, $1$, $2$, and $3$ cm. Note that with a radius of $2$ cm, the peg would be placed about one ring-width away from the expected position, as shown in the supplementary video. The results are shown in Table~\ref{tbl:var}. All controllers were robust to perturbations of $1$ cm, and would often succeed at $2$ cm. Robustness increased slightly when more noise was injected during training, but even controllers trained without noise exhibited considerable resilience to noise. This may be due to the linear-Gaussian controllers themselves injecting noise during sampling, thus improving tolerance to perturbation in future iterations.

\begin{table}[t]
\begin{center}
\footnotesize{
\begin{tabular}{| l | l | l | l | l |}
\hline
\multicolumn{1}{|l|}{lego block} & \multicolumn{4}{|c|}{test perturbation} \\
\hline
training perturbation & 0 cm & 1 cm & 2 cm & 3 cm \\
\hline
0 cm & 5/5 & 5/5 & 3/5 & 2/5 \\
1 cm & 5/5 & 5/5 & 3/5 & 2/5 \\
2 cm & 5/5 & 5/5 & 5/5 & 3/5 \\
kinematic baseline & 5/5 & 0/5 & 0/5 & 0/5 \\
\hline
\hline
\multicolumn{1}{|l|}{ring on peg} & \multicolumn{4}{|c|}{test perturbation} \\
\hline
training perturbation & 0 cm & 1 cm & 2 cm & 3 cm \\
\hline
0 cm & 5/5 & 5/5 & 0/5 & 0/5 \\
1 cm & 5/5 & 5/5 & 3/5 & 0/5 \\
2 cm & 5/5 & 5/5 & 3/5 & 0/5 \\
kinematic baseline & 5/5 & 3/5 & 0/5 & 0/5 \\
\hline
\end{tabular}
}
\end{center}
\vspace{-0.1in}
\caption{Success rates of linear-Gaussian controllers under target object perturbation. Training with larger perturbations improved robustness at test time.
\label{tbl:var}
}
\vspace{-0.2in}
\end{table}

For comparison, we evaluated a kinematic baseline for each perturbation level, which planned a straight path in task space from a point 5 cm above the target to the expected (unperturbed) target location. As shown in Table~\ref{tbl:var}, this baseline was only able to place the lego block in the absence of perturbations. The rounded top of the peg provided a slightly easier condition for the baseline, with occasional successes at higher perturbation levels. However, our controllers outperformed the baseline by a wide margin.

\subsection{Neural Network Controllers}
\label{sec:polresults}

\begin{figure}[t]
\setlength{\unitlength}{0.5\columnwidth}
\vspace{-0.2in}
\begin{picture}(1.99,0.95) \linethickness{0.5pt}

\put(0.0,0.0){\includegraphics[height=0.45\columnwidth]{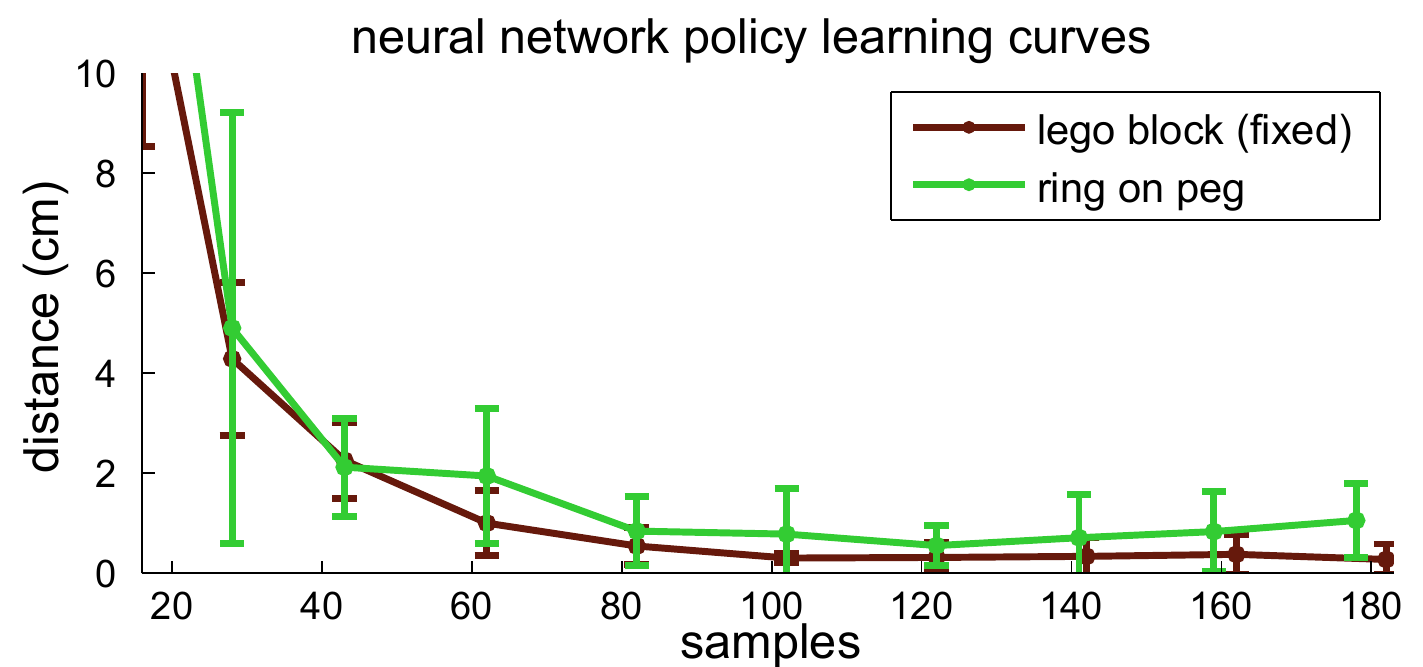}}

\end{picture}
\caption{Distance to target per iteration during neural network training. The samples are divided across four training trajectories. Distance is measured for the individual trajectories only, since the neural network is not run on the robot during training. For performance of the final trained neural network, see Table~\ref{tbl:gen}.
\label{fig:gpsresults}
\vspace{-0.15in}
}
\end{figure}


To demonstrate generalization, we trained two neural network policies using the guided policy search procedure described in Section~\ref{sec:gps}. The first policy was trained on the lego block placement task, with four different target positions located at the corners of a rectangular base of lego blocks. This policy could then be used to place the block at any position on this base. The second policy was trained on the ring task to place the ring on the peg at a range of locations. Figure~\ref{fig:gpsresults} shows learning curves for both tasks. Training neural network policies requires roughly the same number of samples per trajectory as the standard linear-Gaussian method (the graph shows total samples over all trajectories). The additional computation time during training was about 50 minutes, due to the additional cost of optimizing the network. This time could be reduced with a more optimized implementation.

\begin{wrapfigure}{r}{.21\columnwidth}
\vspace{-0.18in}
\hspace{-0.05\columnwidth}\includegraphics[height=0.43\columnwidth]{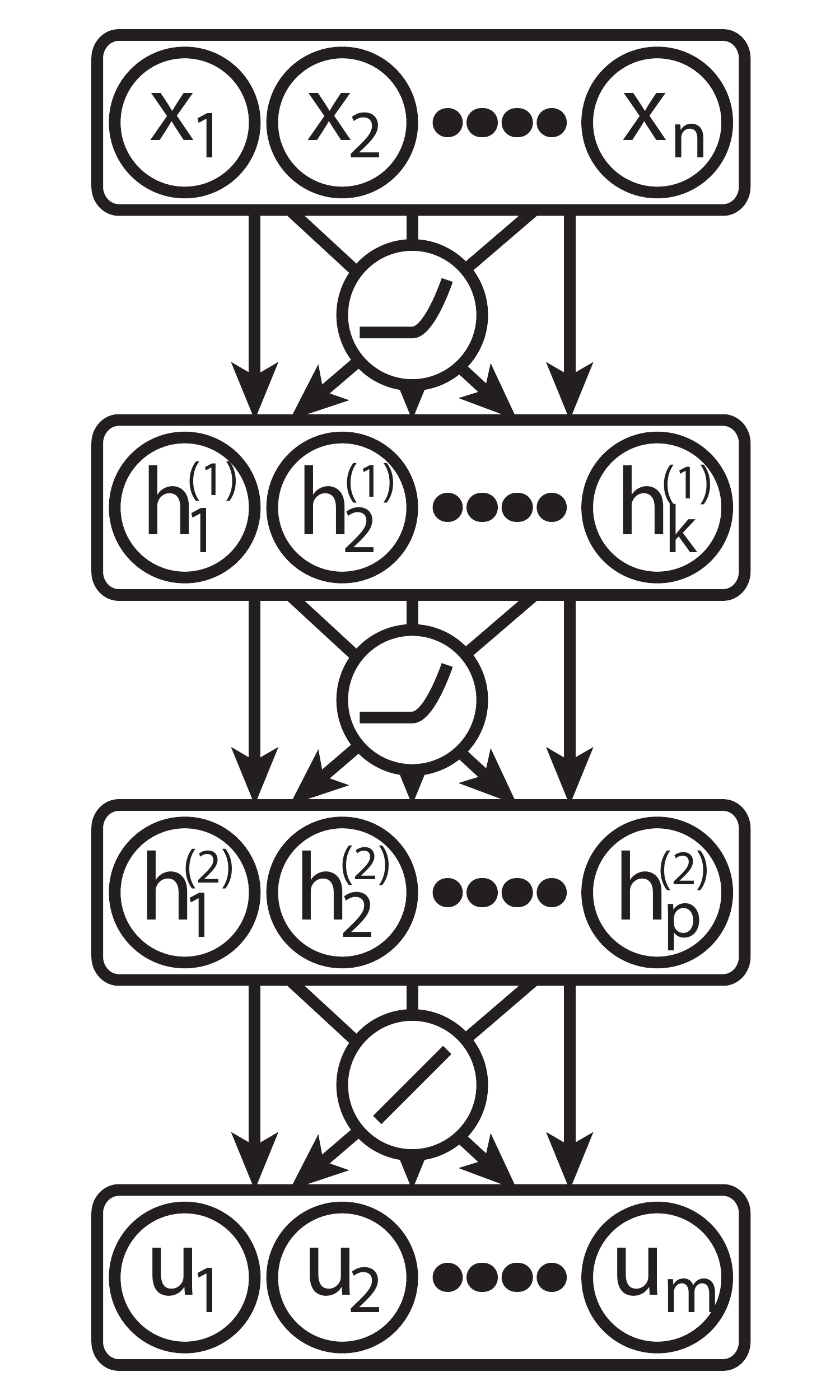}
\vspace{-0.22in}
\end{wrapfigure}
Our neural network policies consisted of two hidden layers with $40$ units each, with soft rectifying nonlinearities between the first two layers, of the form \mbox{$a = \log(z + 1)$}, and linear connections to the output, as shown on the right. Unlike in our previous work \cite{l-edroc-13}, we found that a deeper two-layer architecture was necessary to capture the complexity of the manipulation behaviors.

\begin{table}[b]
\begin{center}
\footnotesize{
\begin{tabular}{| l | l  l  l  l | l  l  l  l  l |}
\hline
\multicolumn{1}{|l|}{lego block} & \multicolumn{4}{|c|}{training positions} & \multicolumn{5}{|c|}{test positions} \\
\hline
position & \#1 & \#2 & \#3 & \#4 & \#1 & \#2 & \#3 & \#4 & \#5 \\
success rate & 5/5 & 4/5 & 5/5 & 3/5 & 5/5 & 5/5 & 5/5 & 5/5 & 5/5 \\
\hline
\hline
\multicolumn{1}{|l|}{ring on peg} & \multicolumn{4}{|c|}{training positions} & \multicolumn{5}{|c|}{test positions} \\
\hline
position & \#1 & \#2 & \#3 & \#4 & \#1 & \#2 & \#3 & \#4 & \#5 \\
success rate & 5/5 & 5/5 & 5/5 & 5/5 & 4/5 & 5/5 & 5/5 & 5/5 & 5/5 \\
\hline
\end{tabular}
}
\end{center}
\vspace{-0.1in}
\caption{Success rates of neural network policies.
\label{tbl:gen}
}
\vspace{-0.1in}
\end{table}

Generalization results for the policies are shown in Table~\ref{tbl:gen}, and videos are presented in the supplementary material. On the lego block task, most of the failures were at the training points, which were at the corners of the rectangular target region. The test points in the interior were somewhat easier to reach, and therefore succeeded more consistently.
On the ring task, the pose of the arm differed significantly between the four training positions, which initially caused the policy to overfit to these positions, effectively learning a classifier for each of the four targets and ignoring the precise target position.
To alleviate this issue, we added noise to the position of the peg during each trial by moving the left gripper which held the peg. This caused the linear-Gaussian controllers to track the target position, and the neural network was able to observe that the target correlated more strongly with success than any particular configuration of joint angles. This allowed it to generalize to all of the test positions.

\section{Discussion}

We presented a method for learning robotic manipulation skills using linear-Gaussian controllers, as well as a technique for training arbitrary nonlinear policies from multiple such controllers using guided policy search. This method can learn linear-Gaussian controllers very quickly with a modest number of real-world trials, can acquire controllers for complex tasks that involve contact and require intricate feedbacks, and can produce controllers that are robust to small perturbations. The nonlinear neural network policies trained with guided policy search can generalize to even larger changes in the task, such as new target locations.


In addition to demonstrating the method on a real robotic platform, we introduced several improvements to guided policy search to make it more practical for robotic applications. We proposed an adaptive step size scheme that speeds up learning, a simple heuristic for adaptively adjusting the number of samples at each iteration, and a way to augment the policy training set with synthetic samples taken from the estimated state-action marginals, which makes it practical to train large neural networks with very few real-world samples.


Our method improves on prior policy search algorithms by learning general-purpose representations with minimal prior knowledge, and by requiring a small amount of interaction time. A central idea in our method is to more tightly control the environment at training time. For example, we might want to handle an arbitrary target for block stacking, but at training time, the same four targets are presented to the robot repeatedly, allowing seperate linear-Gaussian controllers to be learned for each one, and then unified into a single policy.


An additional advantage of guided policy search that was not explored in detail in this work is the ability to provide the policy with a different set of inputs from those available to the linear-Gaussian controllers. For example, the target for an insertion task might be known at training time, but unknown at test time, with only noisy sensors available to the policy. We explored this setting in simulation in our previous work, learning policies that search for the target by ``feeling'' the surface \cite{la-lnnpg-14}. Applying this idea to real-world robotics problems could allow the training of policies that simultaneously perform both perception and control. Learning perception and control jointly could produce policies that compensate for deficiencies in perception with more robust strategies, for example by probing the target before insertion, or dragging the object across the surface to search for the insertion point.



{\bf Acknowledgements} This research was funded in part by DARPA through Young Faculty Award
\#D13AP00046 and by the Army Research Office through the MAST program.

\bibliographystyle{ieeetran}
\bibliography{references}

\end{document}

%% file: defs.tex
\long\def\ignorethis#1{}

\newcommand{\tr}{^\mathrm{T}}
\newcommand{\inv}{^{-1}}

\newcommand{\gauss}{\mathcal{N}}

\newcommand{\vnorm}[1]{\|#1\|}

\newcommand{\costnorm}{r_\ell}

\newcommand{\cluster}{c}

\newcommand{\polwt}{\lambda}
\newcommand{\lagrangian}{\mathcal{L}_{\text{traj}}}
\newcommand{\lagrangiangps}{\mathcal{L}_{\text{GPS}}}
\newcommand{\trajdist}{p}
\newcommand{\policy}{\pi}

\newcommand{\params}{\theta}

\newcommand{\cost}{\ell}
\newcommand{\state}{\mathbf{x}}
\newcommand{\action}{\mathbf{u}}
\newcommand{\hstate}{\hat{\mathbf{x}}}
\newcommand{\haction}{\hat{\mathbf{u}}}

\newcommand{\traj}{\tau}

\newcommand{\ucovar}{\mathbf{C}}

\newcommand{\detpolicy}{g}

\newcommand{\kl}{D_\text{KL}}
\newcommand{\ent}{\mathcal{H}}

\newcommand{\fxt}{f_{\state t}}
\newcommand{\fut}{f_{\action t}}

\newcommand{\fyt}{f_{\state\action t}}

\newcommand{\Qux}{Q_{\action,\state}}

\newcommand{\Qxt}{Q_{\state t}}
\newcommand{\Qut}{Q_{\action t}}
\newcommand{\Qyt}{Q_{\state\action t}}
\newcommand{\Qxxt}{Q_{\state,\state t}}
\newcommand{\Quut}{Q_{\action,\action t}}
\newcommand{\Quxt}{Q_{\action,\state t}}

\newcommand{\Qyyt}{Q_{\state\action,\state\action t}}

\newcommand{\Vxt}{V_{\state t}}
\newcommand{\Vxxt}{V_{\state,\state t}}
\newcommand{\Vxtp}{V_{\state t+1}}
\newcommand{\Vxxtp}{V_{\state,\state t+1}}
\newcommand{\kpol}{\mathbf{k}}
\newcommand{\Kpol}{\mathbf{K}}

\newcommand{\noise}{\mathbf{F}}

\newcommand{\costgradt}{\cost_{\state\action t}}
\newcommand{\costhesst}{\cost_{\state\action,\state\action t}}

\newcommand{\st}{\state_t}
\newcommand{\at}{\action_t}

\newcommand{\points}{\mathbf{p}}

\newcommand{\Penalty}{a}
\newcommand{\Rate}{b}